\pdfoutput=1

\documentclass[11pt]{article}

\usepackage{emnlp2022}

\usepackage{times}
\usepackage{latexsym}

\usepackage[T1]{fontenc}

\usepackage[utf8]{inputenc}

\usepackage{microtype}

\usepackage{booktabs}
\usepackage{graphicx}
\usepackage{amssymb}
\usepackage[titlenumbered,ruled]{algorithm2e}
\usepackage{amsmath}
\usepackage{csquotes}
\usepackage{cleveref}
\usepackage[detect-none]{siunitx}

\title{You Are What You Talk About:\\ Inducing Evaluative Topics for Personality Analysis}

\author{Josip Juki{\'{c}} \quad Iva Vukojevi{\'{c}} \quad Jan {\v{S}}najder\\
University of Zagreb, Faculty of Electrical Engineering and Computing\\
Text Analysis and Knowledge Engineering Lab\\
Unska 3, 10000 Zagreb, Croatia \\
\tt \{josip.jukic, iva.vukojevic, jan.snajder\}@fer.hr
}

\begin{document}
\maketitle

\begin{abstract}
    Expressing attitude or stance toward entities and concepts is an integral part of human behavior and personality. Recently, evaluative language data has become more accessible with social media's rapid growth, enabling large-scale opinion analysis. However, surprisingly little research examines the relationship between personality and evaluative language. To bridge this gap, we introduce the notion of evaluative topics, obtained by applying topic models to pre-filtered evaluative text from social media. We then link evaluative topics to individual text authors to build their evaluative profiles. We apply evaluative profiling to Reddit comments labeled with personality scores and conduct an exploratory study on the relationship between evaluative topics and Big Five personality facets, aiming for a more interpretable, facet-level analysis.
    Finally, we validate our approach by observing correlations consistent with prior research in personality psychology.
\end{abstract}

\section{Introduction}

Sharing opinions has always been rooted in people's daily habits, but nowadays, it has scaled up with a plethora of user-generated texts on social media \citep{lee2012news}. Opinions, as dispositions toward specific entities, can be the key to understanding human behavior. Oftentimes, there is a need to predict sentiment or stance toward a particular target of interest, making opinion analysis useful in marketing research, social and political sciences, and recommender systems, to name a few. Moreover, analyzing how we express our opinions, i.e., the linguistic aspect and the psychological disposition of the opinion holder, is crucial for various downstream tasks, such as personality analysis and mental health prediction.

\begin{figure}[t!]
\centering
\includegraphics[width=\linewidth]{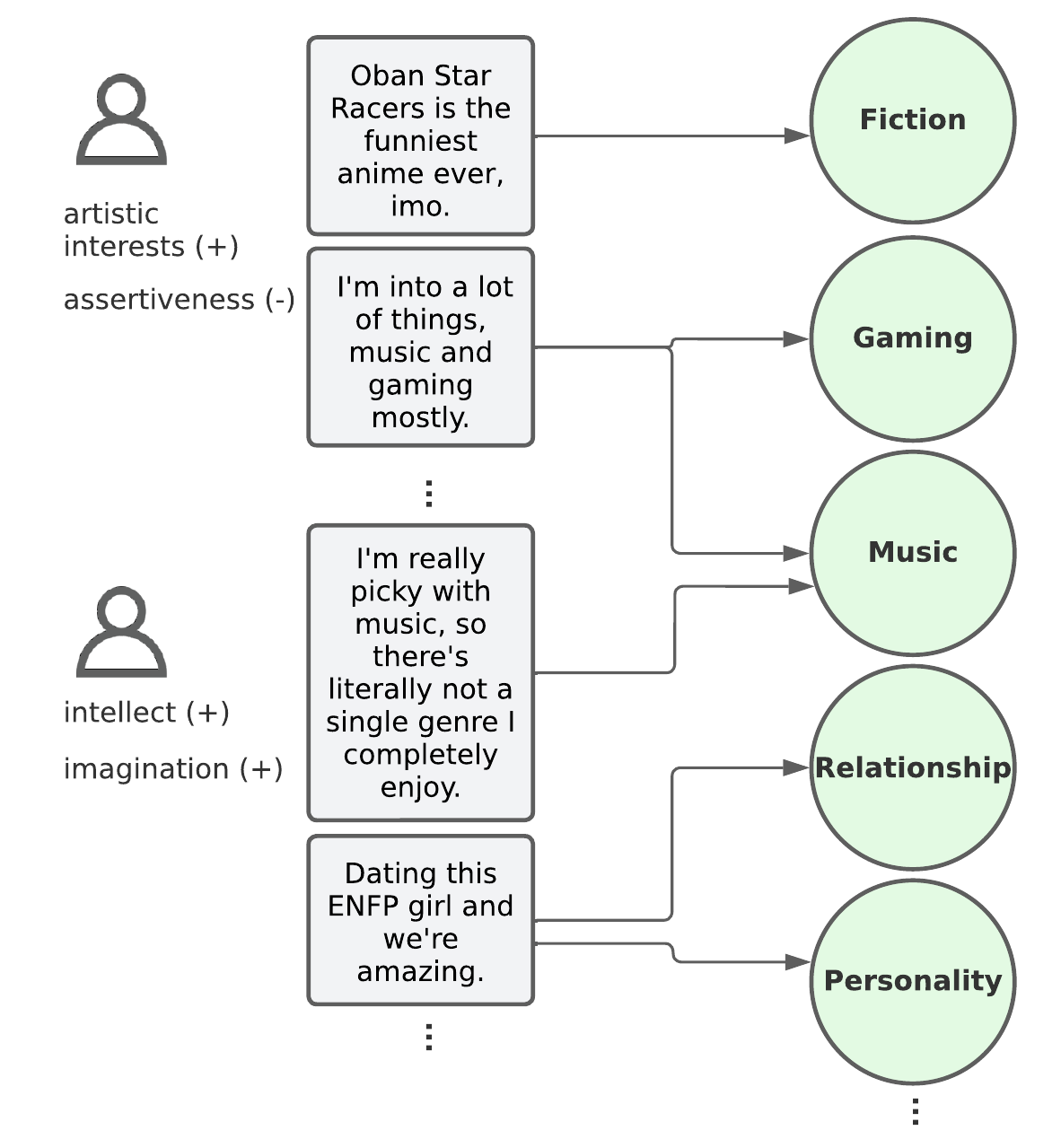}
\caption{An illustrative example of mapping user statements to \emph{evaluative topics}, i.e., topics induced from evaluative texts. Based on the prevalence of such topics, we construct an \emph{evaluative profile} of each author. We then measure the correlation between each author profile and personality facets, such as artistic interests, assertiveness, intellect, and imagination.}
\label{fig:topics}
\end{figure}

In public communication, among other situations, people use linguistic resources known as \textit{evaluative language} \citep{hunston2010corpus} to express their attitudes. \citet{hunston2000evaluation} define evaluation as a cover term for unified phenomena of certainty and goodness/desirability, and evaluative language includes every expression of the speaker or writer’s attitude or stance towards, or feelings about the entities or propositions. As evaluation is associated with psychological disposition \cite{malrieu1999evaluative}, evaluative language can help uncover people's distinguishing qualities, i.e., recurrent behavioral, cognitive, or affective tendencies. Such behavioral differences resulting from interaction with the environment fall within the purview of \textit{personality psychology}, where they are typically examined as \textit{personality traits} -- sets of characteristic patterns of behavior, cognition, and feelings stable over time and across situations \citep{funder2012accurate}. Moreover, personality traits have been shown to correlate with occupational preferences, political or religious tendencies, intelligence, personal interests, and opinions \citep{larson2002interests, ackerman1997intelligence, hyman-1957-exploration}.

To the best of our knowledge, evaluative language has not yet been studied in the context of personality analysis on social media. This can perhaps be attributed to methodological challenges. One of the difficulties in analyzing evaluative language is separating the signal from the noise. Although evaluation in text is pervasive, it remains a challenge to automatically distinguish evaluative from non-evaluative text. Evaluative expressions come in all sort of flavors, with varying amounts of explicit lexical stance markers. Another difficulty is choosing the right level of abstraction both for personality and evaluative language. If the evaluation targets are too specific, it is difficult to relate them to personality. Nevertheless, we can choose the appropriate operationalization of the personality in terms of its breadth, thus aligning it with the abstraction level of the evaluative targets.

In this paper, we study how evaluative language on social media can be used for personality analysis. More specifically, our study aims to explore how the topics people talk about are related to their personality, as illustrated in \Cref{fig:topics}. To this end, we introduce the notion of an \emph{evaluative topic}, which is a topic constructed from evaluative text. To address the signal-noise problem in detecting evaluative text, we develop an iterative filtering technique based on detecting evaluative expressions and paraphrase mining. We then leverage topic models to induce evaluative topics from prefiltered evaluative text. Aiming to maximize coherence and diversity, we test several topic model families, including probabilistic, decompositional, and neural topic models. Finally, we link evaluative topics to individual text authors to construct authors' \emph{evaluative profiles}.

In the experimental part, we investigate the relationship between evaluation and personality. Our focus is on the \textit{Big Five} model of personality \citep{goldberg-1981-language}, which includes broad personality traits and their more specific (and arguably more interpretable) aspects called \textit{facets}.
We present a correlation study of evaluative topics and Big Five facets on comments from Reddit, one of the most popular discussion websites, which covers an astoundingly diverse range of topics while preserving user anonymity. We find correlations between evaluation and personality that are consistent with prior research in personality psychology. Finally, we provide empirical evidence that evaluative statements have a stronger association with personality than non-evaluative expressions.

In summary, our contribution is threefold: we (1) develop an approach for evaluative author profiling based on topic models with evaluative filtering, (2) apply evaluative profiling on Reddit comments, and (3) study how evaluative author profiles of Reddit users correlate with Big Five personality facets.
    
The rest of the paper is structured as follows. Section \ref{sec:rw} lays out the background and related work. We describe our technique for filtering evaluative text along with topic modeling in Section \ref{sec:eval}. We conclude the section by presenting the idea of evaluative author profiles. We describe our experiments and results in Section \ref{sec:experiments}. Finally, Section \ref{sec:conclusion} overlays the main conclusions and takeaways of the paper.

\section{Background and Related Work}
\label{sec:rw}

\paragraph{Evaluative language.}
In linguistics, studies of evaluative language are embedded in functional approaches to language that aim to understand and explain linguistic structures in terms of the semantic and communicative functions of language and which assume that its primary function is to be a vehicle for social interaction \citep{allan2007western}. Most research in evaluative language covers only a particular niche of evaluation, such as subjectivity \cite{wiebe-etal-2004-learning} and stance \cite{kiesling-etal-2018-interactional}. For instance, \citet{biberfinegan} describe lexical and grammatical markers of stance, i.e., categorizations along the continuum of epistemic and attitudinal meanings, developing groups of markers for people's attitude, affect, and assessment. More recently, \citet{pavalanathan-etal-2017-multidimensional} compiled a lexicon of stance markers on Reddit data. On the other hand, because sentiment analysis is ubiquitous in NLP, evaluative language has been viewed through the prism of sentiment \cite{benamara-etal-2017-evaluative, bo-lee-2008-opinion}.
In our work, we attempt to combine several aspects of evaluative language, namely sentiment, stance, and opinion markers.

\paragraph{Target awareness.}
Given the nature of expressing opinions, evaluation is directed toward a particular target. Among others, sentiment-based approaches to evaluation have been prevalent, which led to the establishment of aspect-based sentiment analysis (ABSA), an NLP task that combines aspect extraction with sentiment classification \citep{pontiki-etal-2014-semeval}. Recent research in ABSA has evolved from traditional perspectives, e.g., using conditional random fields for aspect extraction as a sequence labeling task \cite{toh-wang-2014-dlirec}, to neural-based approaches \cite{he-etal-2017-unsupervised, ijcai2019-712, hoang-etal-2019-aspect, tulkens-van-cranenburgh-2020-embarrassingly}. However, ABSA systems are better suited for smaller domains. Besides ABSA, there have been target-based approaches in stance classification \cite{du2017stance} and topic-dependent argument classification \cite{reimers-etal-2019-classification}, among others. As we are trying to cover diverse domains, we opt for a topic-level approach to targets.

\paragraph{NLP and personality.}
NLP has been used extensively for personality analysis, starting with essay analysis \cite{pennebaker-king-2000-linguistic} and followed by social network research \cite{park-etal-2014-automatic, schwartz-etal-2013-personality}. The more recent studies were mainly conducted with Facebook data. For example,  \citet{kulkarni2018latent} used trait-level factor analysis for the Big 5 personality system on Facebook. However, as far as we know, there is no research that leverages NLP to analyze personality in the context of evaluative language. Moreover, there is little research in NLP that covers facet-level personality analysis.

\section{Methodology}
\label{sec:eval}

We propose a three-step process from raw text to evaluative profiles. We start with evaluative filtering to extract evaluative text and then construct topics from the filtered dataset. Finally, we use the developed evaluative topics to create the corresponding evaluative profiles.

\subsection{Dataset}

We used \textsc{Pandora}, a dataset with personality scores for Reddit users, extracted from self-reported personality questionnaire results \citep{gjurkovic2021pandora}. In total, there are 1,608 users with self-reported Big Five scores. These users have written a total of 1.3M comments, consisting of 14.3M sentences. Additionally, 127 users self-reported their questionnaire scores for the NEO PI-R facets.\footnote{\href{https://ipip.ori.org/newNEO\_FacetsTable.htm}{https://ipip.ori.org/newNEO\_FacetsTable.htm}} NEO PI-R provides information on six facets of each Big Five personality trait. A facet is a specific and unique aspect of a broader personality trait. For example, \textit{anxiety} and \textit{depression} are facets of \textit{neuroticism}, while \textit{friendliness} and \textit{gregariousness} are extraversion facets. After we applied preprocessing, which included segmenting the comments into sentences, we ended up with 6.5M sentences. We filtered out non-English comments since we used models pre-trained on English texts. We attach more details on the preprocessing step in Appendix~\ref{subsec:A_dataset}.

\subsection{Evaluative filtering}

Although social media abounds with evaluative language, it is challenging to automatically distinguish evaluative from non-evaluative text. To filter out non-evaluative expressions, we collected comments with evaluative markers -- lexical or grammatical language forms that express the act of evaluation. We combined off-the-shelf sentiment analysis tools with opinion and stance lexicons to cover different aspects of evaluative language. On top of that, we searched for sentences that contain \textit{evaluative patterns}, i.e., phrases that express opinions.

To extract sentiment-laden sentences, we used VADER \citep{hutto-gilbert-2014-vader}, where we summed the positive and negative VADER scores to determine the overall sentiment score. We used the opinion lexicon devised by \citet{hu2004mining} and the lexicon of stance markers \citet{pavalanathan-etal-2017-multidimensional}. Inspired by \citet{hunston-thompson-2000-evaluation}, we compiled a list of evaluative patterns in the form of regular expressions that we matched across the dataset. These include phrases such as \textit{I like/hate/believe/support}, \textit{personally}, \textit{in my (honest) opinion}, etc. Along with variations of the aforementioned phrases, we extended the regular expressions to support negations and modifiers, as well as the vernacular language used in the online community (e.g., \textit{IMO} -- \textit{in my opinion}, \textit{FMPOV} -- \textit{from my point of view}). To collect matched sentences, we set a constraint that a given match must exceed the thresholds for sentiment, opinion, and stance scores. We took the intersection of matched sentences in the 50th percentile for each category except sentiment, for which we set an upper bound. This is because we found that sentences with extreme sentiment scores often do not have an explicit target, and the target cannot be induced without additional context.\footnote{We decided not to use sentences above the 90th percentile in sentiment, which typically indicates sentences expressing emotions with highly implicit targets (e.g., a sentence with many emojis).}

After applying evaluative filtering on \textsc{Pandora}, we obtained 29k sentences, focusing on precision as we used strict pattern matching with high evaluative scores. To improve recall, we developed quasi-snowballing (QSB), a simple paraphrase mining technique.\footnote{The code for quasi-snowballing is available at the link: \href{https://github.com/josipjukic/quasi-snowballing}{https://github.com/josipjukic/quasi-snowballing}} QSB is an iterative procedure that starts with a seed set of evaluative expressions and extends it with similar statements. We employed sentence transformers\footnote{We chose \textit{all-mpnet-base-v2} from \href{https://www.sbert.net/}{https://www.sbert.net/}} \citep{reimers-gurevych-2019-sentence} to compute contextualized representations and use these representations to detect paraphrases. We initialized QSB with the filtered evaluative sentences as the seed set. Afterward, we used cosine similarity as a criterion to extract similar sentences according to the similarity threshold $t_{sim}$, which is adjusted at each step by the similarity growth factor $\gamma$. Inspired by simulated annealing, the factor $\gamma$ increases the threshold exponentially to achieve easier convergence as the whole set of sentences grows with each iteration. At the end of each iteration, we augmented the old seed set with the newly obtained matches. The process stops when there are no more candidates. Using QSB, we obtained 310k sentences with evaluative markers (details in Appendix~\ref{subsec:A_filtering}).

QSB mines expressions that can differ in target, polarity, or intensity. Since we use a high similarity threshold, in most cases only one of the above three components will be different in a paraphrase match. Multiple iterations of paraphrase mining can evolve the original sentence, resulting in more lenient matches overall, as shown in Table~\ref{tab:snow}.

\begin{table*}
\centering
\small
\begin{tabular}{p{0.4\linewidth}  p{0.4\linewidth} c}
\toprule
Original hit & Paraphrase match & Similarity\\
\midrule
Age is inexcusable to lie about \textbf{IMO}. & I'm not a fan of lying about age. & .86 \\
\textbf{I support} the death penalty, but I would never label myself as pro-life. & I'm pro-life in the sense that I would rather not have people abort later term when it could be reasonably considered a person... & .77 \\
I'm torn on this one, because \textbf{I support} trans folks 100\% and \textbf{I believe} that a trans woman is a woman, end of story, and same for a man. & I'm supportive of transgender people transitioning and being legally treated as someone of the opposite sex. & .78 \\
\bottomrule
\end{tabular}
\caption{Examples of QSB paraphrase matching. The first column represents the comments that are matched to regular expressions with matched text shown in bold. The second column lists matched paraphrases.}
\label{tab:snow}
\end{table*}

\subsection{Evaluative topics}

In the second step, we applied topic modeling to the pre-filtered evaluative comments to obtain evaluative topics. We adopted the definition of a topic as a distribution over a fixed vocabulary of terms \citep{blei-etal-2009-topic}, and further defined an \textit{evaluative topic} as a topic built from target-specific opinions derived from evaluative language. To produce evaluative topics, we experimented with traditional probabilistic models as well as neural-based architectures. Furthermore, since we are dealing with Reddit comments, which tend to be short, we considered several topic modeling approaches for short texts.

\paragraph{LDA.} From the family of traditional topic models, we opted for the latent Dirichlet allocation (LDA), a well-established probabilistic topic model \citep{blei2003latent}, which also serves as a strong baseline. Considering that LDA deals poorly with short texts due to the data sparsity problem \citep{hong2010empirical}, following \citet{zuo-etal-2016-topic}, we grouped comments into larger documents. For each author, we created pseudo-documents grouped by subreddit, i.e., a forum dedicated to a particular topic on Reddit.

\paragraph{BTM.} We tested the biterm topic model (BTM), which is primarily designed for short texts \citep{cheng2014btm}. We fed the model with comment-level text, where we filtered out non-evaluative sentences from the comment.

\paragraph{ABAE.} We tried out a neural-based architecture in \textit{Attention-based Aspect Extraction} (ABAE), an autoencoder proposed by  \citet{he-etal-2017-unsupervised}. We include this model as it is designed specifically for aspect clustering, which we expect to align well with our objective of building evaluative topics. ABAE exploits word embeddings to improve coherence and attempts to group together words that appear in similar contexts. Furthermore, ABAE uses the attention mechanism to reduce the relative importance of irrelevant words during training. We adopted ABAE but made a few changes to the training procedure. First, we trained custom Word2Vec embeddings \citep{mikolov2013efficient} on \textsc{Pandora}. We also modified the reconstruction loss function from the ordinary dot product to match cosine similarity with values in the $[-1,1]$ interval, which improved the learning stability. We fed the model with segmented evaluative sentences to learn the topics.

\paragraph{CTM.} Finally, we experimented with the combined topic model (CTM) from \cite{bianchi-etal-2021-pre}, which has been shown to increase coherence compared to other topic models. CTM is a blend of a variational autoencoder and embeddings from a sentence transformer.\footnote{We used \textit{paraphrase-distilroberta-base-v2} to produce contextual embeddings.}

\paragraph{}
Topic models are usually evaluated by means of topic coherence. However, this method suffers from a validation gap, i.e., automated coherence is not validated by human experimentation \citep{holye-etal-2021-automated}. To mitigate this problem, we count token co-occurrences on the whole dataset and not only on the training data, which has been shown to have a stronger association with human judgment \citep{ding-etal-2018-coherence}. Specifically, we used the normalized pointwise mutual information (NPMI) to evaluate the coherence of the induced topics. For evaluating diversity, we adopted the metric proposed by  \citet{bianchi-etal-2021-pre}, defined as the reciprocal of the standard RBO \cite{terragni-etal-2021-natural}.\footnote{RBO uses a weighted ranking to estimate the disjointedness between the top 10 words for each topic.} We used NPMI and IRBO as two criteria of a multi-objective optimization procedure. We determined the Pareto front of the trained models, i.e., the set of non-dominated models, from which we chose the model with the smallest number of topics (Table \ref{tab:topic}).

\begin{table}
\centering
\small
\begin{tabular}{lrrc}
\toprule
& NPMI & IRBO & \#\,topics
\\
\midrule 
LDA & $-.1413$ & $.9905$ & $20$ \\
BTM & $-.2159$ & $.8241$ & $30$ \\
ABAE & $-.0521$ & $.9833$ & $20$ \\
CTM & $\mathbf{.0628}$ & $.9981$ & $20$ \\
\bottomrule
\end{tabular}
\caption{Topic modeling evaluation results.
NPMI is a coherence score in $[-1, 1]$ range, with $-1$ indicating that the topic representative terms occurred together, $0$ indicating that the term occurrences were independent of each other, and $1$ indicating that the terms co-occurred perfectly with each other. IRBO measures topic diversity with $1$ for completely different and $0$ for identical topics. We repeated each experiment $10$ times with different seeds. Shown in bold is the score that is significantly better than the scores for the rest of the models (independent Wilcoxon test for each model pair with $p < .01$, adjusted for family-wise error rate with the Holm-Bonferroni method).}
\label{tab:topic}
\end{table}

\subsection{Evaluative author profiles}

\def\d{
\begin{bmatrix}
    c_1 c_2 \hdots c_K\\
\end{bmatrix}}

\def\Uext{
\begin{bmatrix}
    u_{1}^{(1)} u_{2}^{(1)} \hdots u_{K}^{(1)} \\
    u_{1}^{(2)} u_{2}^{(2)} \hdots u_{K}^{(2)} \\
    \vdots \\
    u_{1}^{(N)} u_{2}^{(N)} \hdots u_{K}^{(N)} \\
\end{bmatrix}}

\def\U{
\begin{bmatrix}
    \mathbf{u}^{(1)}
    \mathbf{u}^{(2)}
    \hdots
    \mathbf{u}^{(N)}
\end{bmatrix}}

\def\Sext{
\begin{bmatrix}
    s_{1}^{(1)} s_{2}^{(1)} \hdots s_{K}^{(1)} \\
    s_{1}^{(2)} s_{2}^{(2)} \hdots s_{K}^{(2)} \\
    \vdots \\
    s_{1}^{(N)} s_{2}^{(N)} \hdots s_{K}^{(N)} \\
\end{bmatrix}}

\def\S{
\begin{bmatrix}
    \mathbf{s}^{(1)}
    \mathbf{s}^{(2)}
    \hdots
    \mathbf{s}^{(N)}
\end{bmatrix}}

\def\V{
\begin{bmatrix}
    \mathbf{v}^{(1)}
    \mathbf{v}^{(2)}
    \hdots
    \mathbf{v}^{(N)}
\end{bmatrix}}

In the third step, we leverage evaluative topics to produce evaluative author profiles. Specifically, we describe each user in terms of topic prevalence, where user text is distributed as a sentiment-weighted mixture of topics across different targets. Each user is thus assigned the average topic prevalence. We compute the average topic distributions for the entire collection of user sentences, where a given sentence contributes with a value from the $[0,1]$ interval for each topic.

To formalize our procedure, we begin by defining a topic distribution for a specific document $\mathbf{d} = \d^\mathsf{T}$, where $\mathbf{d}$ represents a document (e.g., sentence, comment), and $c_k$ is the corresponding mixture component for the $k$-th topic.
We aggregate the values of the user's evaluative sentences to compute the $n$-th user's components for each topic and concatenate the aggregations into the vector $\mathbf{u}^{(n)}$:
    \[
        \mathbf{u}^{(n)} = \frac{1}{N_n} \sum_{i=1}^{N_n} \mathbf{d}^{(n, i)} \text{,}
    \]
where $N_n$ is the number of the $n$-th user's documents and $\mathbf{d}^{(n, i)}$ is the $i$-th document of the $n$-th user.
Moreover, we incorporate the sentiment intensity information to obtain a sentiment-enhanced representation $\mathbf{v}^{(n)}$:
    \[
        \mathbf{v}^{(n)} = \frac{1}{N_n} \sum_{i=1}^{N_n} s^{(n, i)} \mathbf{d}^{(n, i)} \text{,}
    \]
where $s^{(n, i)}$ is the sentiment intensity for the $i$-th document of the $n$-th user calculated as the sum of the positive and negative VADER scores. 

The use of sentiment intensity in lieu of polarity deserves further explanation. When considering sentiment polarity, we can discern two types: the user's sentiment polarity for a given statement and the implicit polarity \citep{russo-etal-2015-semeval} of the topic itself. This gives rise to a number of different ways in which sentiment information can be incorporated into evaluative representations. However, the inclusion of sentiment polarity makes it difficult to distinguish whether the topic sentiment is driven primarily by implicit or explicit polarity. We sought parsimony, primarily to facilitate explainability, and thus chose to use only sentiment intensity. Future work may look into combining both types of sentiment.
\section{Experiments}
\label{sec:experiments}

Our investigation of the association between evaluative language and personality proceeded in two steps. We first built evaluative author profiles from Reddit comments from \textsc{Pandora}, as described in the previous section. To build evaluative topics, we used the CTM model on the filtered evaluative sentences, as CTM yielded the best result in terms of coherence and diversity scores. We induced $20$ evaluative topics, as per the results of the optimization process. \Cref{tab:cluster} shows a sample of the topics alongside manually assigned labels.

In the next step, we carried out a correlation study on the relationship of Big Five personality facets and evaluative topics. We adopted the Revised NEO Personality Inventory (NEO PI-R), designed to measure the Big Five personality traits: openness to experience, conscientiousness, extraversion, agreeableness, and neuroticism \cite{costa-etal-1995-domains}. In the first part of the analysis, we examined individual correlations between particular topics and facets. In the second part, we used canonical correlation analysis (CCA) to explore common associations between the entire set of topics on the one side and the set of all Big Five facets on the other. We chose to focus on personality facets rather than personality traits, hypothesizing that the abstraction level of facets aligns better with the granularity of induced topics. In that sense, the choice of facets supports the Brunswik symmetry principle, which stipulates that analyzed constructs (in our case, personality and topics) must have similar levels of generality \cite{wittmann2012principles}.

\subsection{Pairwise correlations}

\begin{figure*}[t!]
\centering
\includegraphics[width=\linewidth]{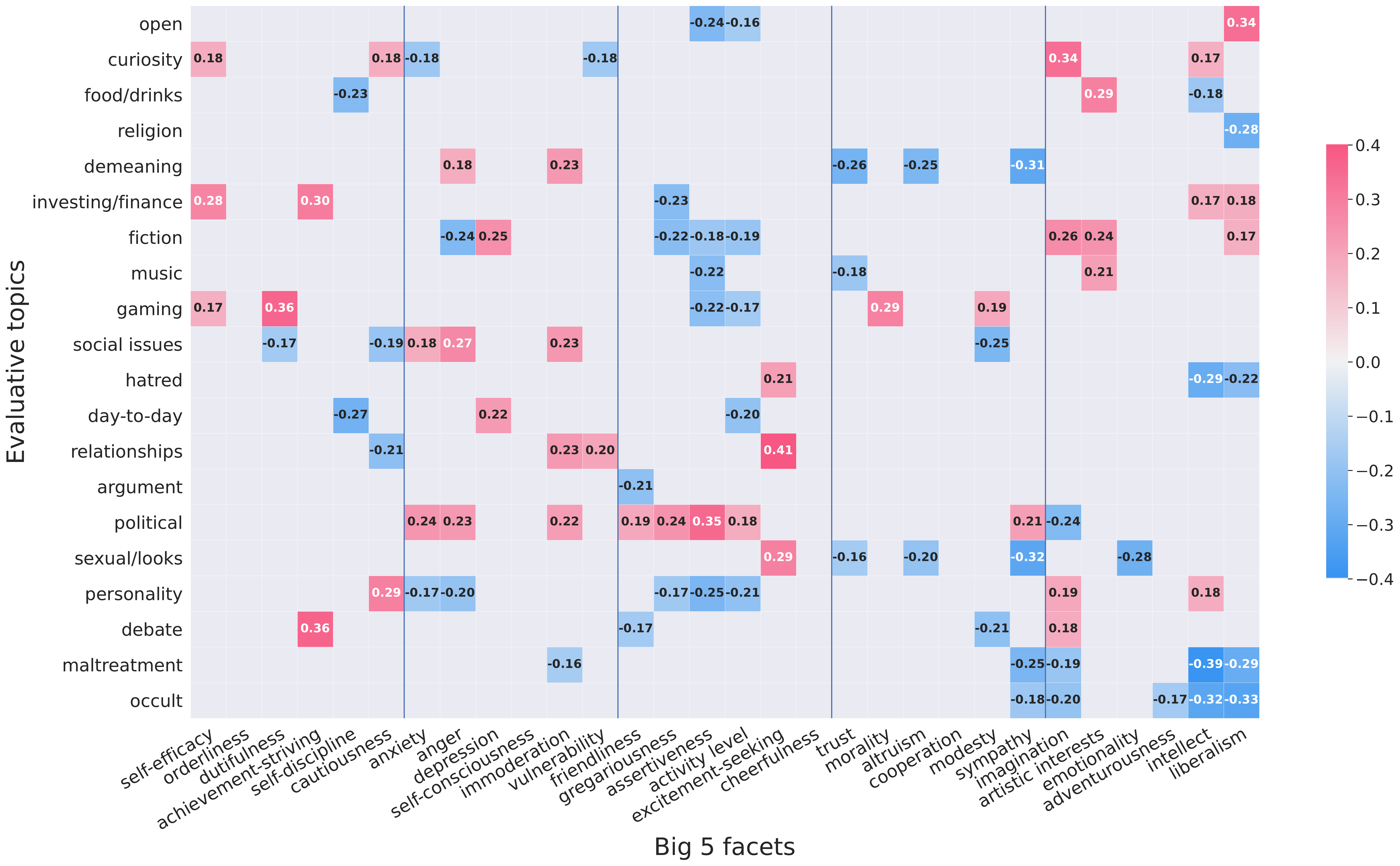}
\caption{Pairwise partial correlations between Big 5 facets (x-axis) and evaluative topics (y-axis) with control for gender.  We show only significant correlations ($p < .01$), determined using Fisher's z-transformation of correlations and corrected for false discovery rate with the Benjamini-Hochberg method.}
\label{fig:corr}
\end{figure*}

We calculated partial pairwise correlations between evaluative author profiles and Big Five facets with control for gender as a possible confounder. For gender, we used the values provided with the \textsc{Pandora} dataset, which surpassed the $F_1$ score of $.90$ \citep{gjurkovic2021pandora}.\footnote{The gender prediction model of \citet{gjurkovic2021pandora} is a binary classifier trained only on a subset of users that self-reported one of the two binary gender categories. Consequently, its predictions will be incorrect for users with non-binary gender. However, we presume that this, along with the fact that even for binary cases, predictions will not always be correct, has only a limited effect on correlation estimates partialed out for gender.}
\Cref{fig:corr} shows significant correlations corrected for false discovery rate with the Benjamini-Hochberg method. We observe numerous small correlations and even moderate correlations ($> .4)$ for some topic-facet pairs \citep{bosco-et-al-2015-correlational}. This is surprising, given that the average correlation in questionnaire-based studies of individual differences is $.19$ \cite{gignac2016effect} and that text data are noisier than questionnaire data.

To assess the validity of our results, we looked into personality psychology literature for reference. According to psychological research, facets of openness to experience are expected to correlate positively with \textit{curiosity}, \textit{food/drinks}, \textit{fiction}, \textit{music}, \textit{gaming}, \textit{political}, \textit{personality}, \textit{debate}, and \textit{occult} topics, while they correlate negatively with the \textit{religion} topic \cite{stewart2022finer, ozer2006personality, soto2019replicable, intiful2019exploring, skimina2021traits, church2008prediction, marshall2015big, chauvin2021individual}. As \Cref{fig:corr} shows, the majority of our results are consistent with mentioned prior research. Three listed associations that we do not significantly support are positive correlations with \textit{debate} (hypothesized direction, but not significant), \textit{gaming} (hypothesized direction, but not significant), and \textit{occult} (opposite sign). However, we note that in most of the referenced psychological literature, effects were analyzed at the trait level rather than the facet level.

\begin{table*}
\centering
\small
\begin{tabular}{ll}
\toprule
Label & Representants \\
\midrule 
open & people, lot, free, good, permit, time, going, open, big, trying \\
food/drinks & taste, meat, smell, beer, texture, flavour, savory, drink, palate, sweet \\
religion & testimony, believe, truth, god, witness, primordial, bible, church, jesus, earth \\
demeaning & shit, tell, talking, strangers, rude, stupid, bitch, upset, weird, shitty \\
investing/finance & recommend, pack, vouch, opt, buying, invest, fund, ticket, stock, dealt \\
fiction & watch, character, read, movies, story, books, thought, cool, new, anime \\
music & songs, album, pop, favourite, music, voice, listen, lyrics, track, sound \\
gaming & play, game, level, team, damage, skill, hit, gear, combat, dungeons \\
social issues & society, human, religion, culture, rights, moral, white, argument, victimhood, black \\
hatred & despise, hate, passion, stand, fucking, goddamn, nerd, mad, smug, ads \\
day-to-day & day, work, hours, week, spend, food, money, sleep, home, eat \\
relationships & relationship, woman, dating, child, partner, man, sex, ex, gay, alimony \\
argument & opinion, salt, grain, sanctimonious, worthless, controversial, assessment, disclaimer, 180, nuanced \\
political & government, party, vote, trump, support, country, state, speaker, system, tax \\
sexual/looks & hot, hair, sexy, gross, facial, body, attractive, wear, porn, dimorphism \\
personality & type, personality, mbti, emotions, test, cognitive, learning, ideas, brain, jung \\
debate & understand, discussion, post, saying, person, wrong, think, point, internet, debate \\
maltreatment & despised, hated, school, dropped, sucked, harass, refuses, skipped, hood, beaten \\
occult & fate, chime, cat, spirits, guides, luck, desperately, talisman, invite, mirror \\
\bottomrule
\end{tabular}
\caption{Evaluative topics produced by the CTM. The topic labels shown in first column were manually assigned as the most frequent label among five annotators. The second columns shows the top $10$ words for each topic.}
\label{tab:cluster}
\end{table*}

\subsection{Canonical correlation analysis}
As facets are inherently intertwined, we can gain deeper insight from analyzing the facets jointly with respect to the set of evaluative topics. To this end, we employed canonical correlation analysis (CCA). The goal of CCA is to find a linear combination of evaluative profiles on the one side and facets on the other, with the goal of maximizing the correlation between the newly created canonical variables. CCA assumes that both variable sets have multivariate normal distributions, which we confirmed with the Henze-Zirkler test \citep{henze-zirkler}.\footnote{Normality was not rejected in both cases, where $p=.62$ and $p=.51$ for evaluative and personality variables, respectively.} We computed $20$ canonical dimensions and sorted them in descending order of correlation magnitude.\footnote{In general, the number of canonical dimensions is equal to the number of variables in the set with fewer variables.} We found statistically significant correlations for the first three pairs of canonical variates with Wilks' lambda test.\footnote{We observed significant correlation ($p < .01$) for the following ranges: $\numrange{1}{20}$, $\numrange{2}{20}$, and $\numrange{3}{20}$.}

\begin{figure}[t!]
\centering
\includegraphics[width=\linewidth]{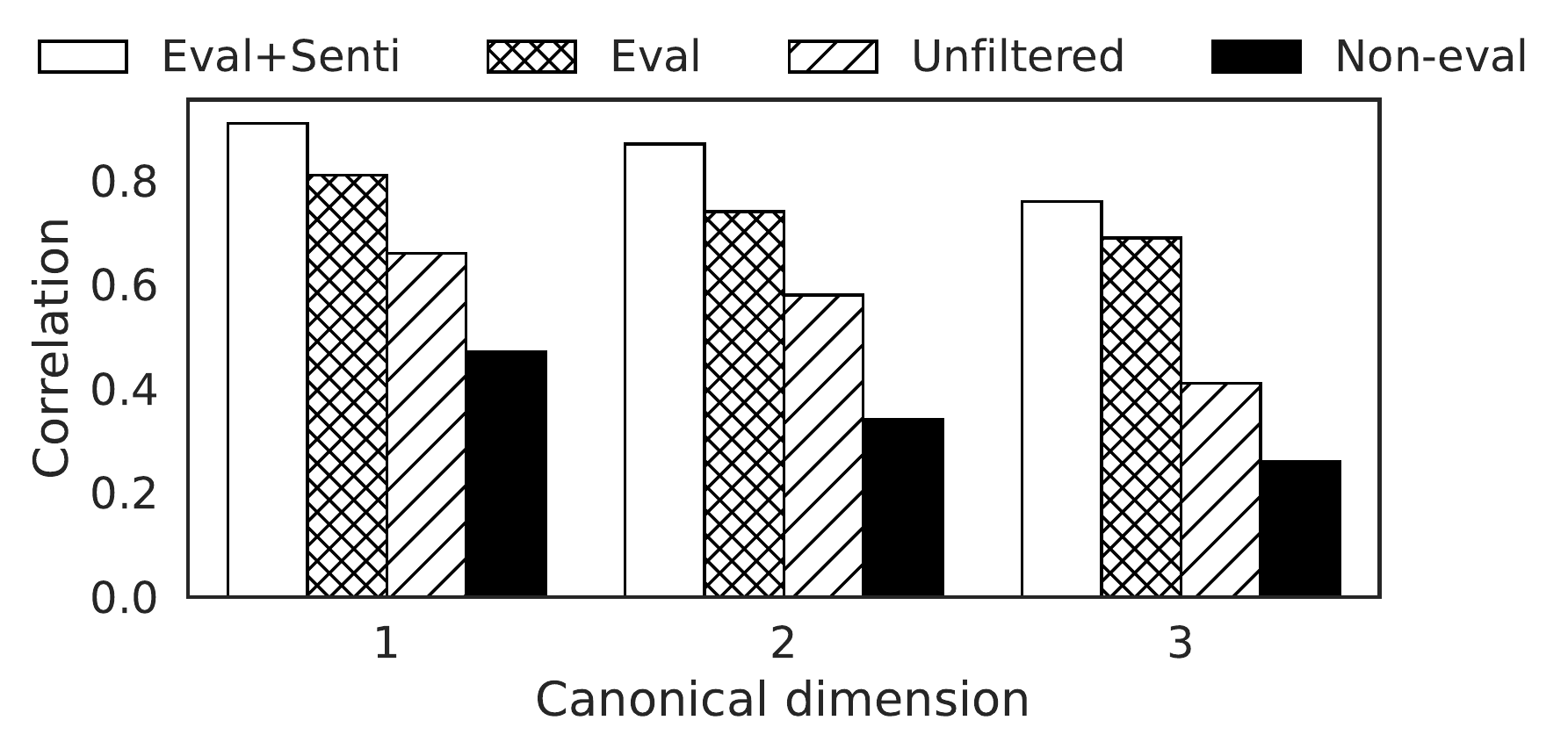}
\caption{Canonical correlations of the first three canonical pairs ($p < .01$) estimated on four data subsets: original text (\textit{Unfiltered}), text obtained with evaluative filtering (\textit{Eval}), \text{Eval} with sentiment information included when calculating the evaluative profiles (\textit{Eval+Senti}), and non-evaluative text (\textit{Non-eval}) obtained as  the set difference between \textit{Non-filtered} and \textit{Eval} sets.}
\label{fig:ccacomp}
\end{figure}

An additional question we wanted to answer is whether personality analysis benefits from evaluative topics (constructed from pre-filtered evaluative text) as opposed to ordinary topics (constructed from all text). To this end, we applied CCA to three different data subsets: (1) unfiltered text, (2) text obtained with evaluative filtering, and (3) non-evaluative text, obtained as the difference between the first and the second set. In addition, we ran CCA on (4) evaluative text with sentiment intensity in order to investigate whether evaluative profiles benefit from sentiment information. \Cref{fig:ccacomp} shows the canonical correlations of the first three canonical pairs on the four data subsets. Higher correlations computed by CCA indicate stronger associations between the two sets of variables. Thus, under the assumption that evaluative profiles and personality are associated, obtaining higher correlations supports construct validity of evaluative profiles as construed by our model. Construct validity, in this case, concerns the extent to which an evaluative topic accurately assesses what it is designed for. With this in mind, two main observations emerge from the results: (1) evaluative pre-filtering seems to be more apt than using unfiltered data for establishing an association with Big 5 personality and (2) sentiment information amplifies the correlations.

\begin{figure}[t!]
\centering
\includegraphics[width=\linewidth]{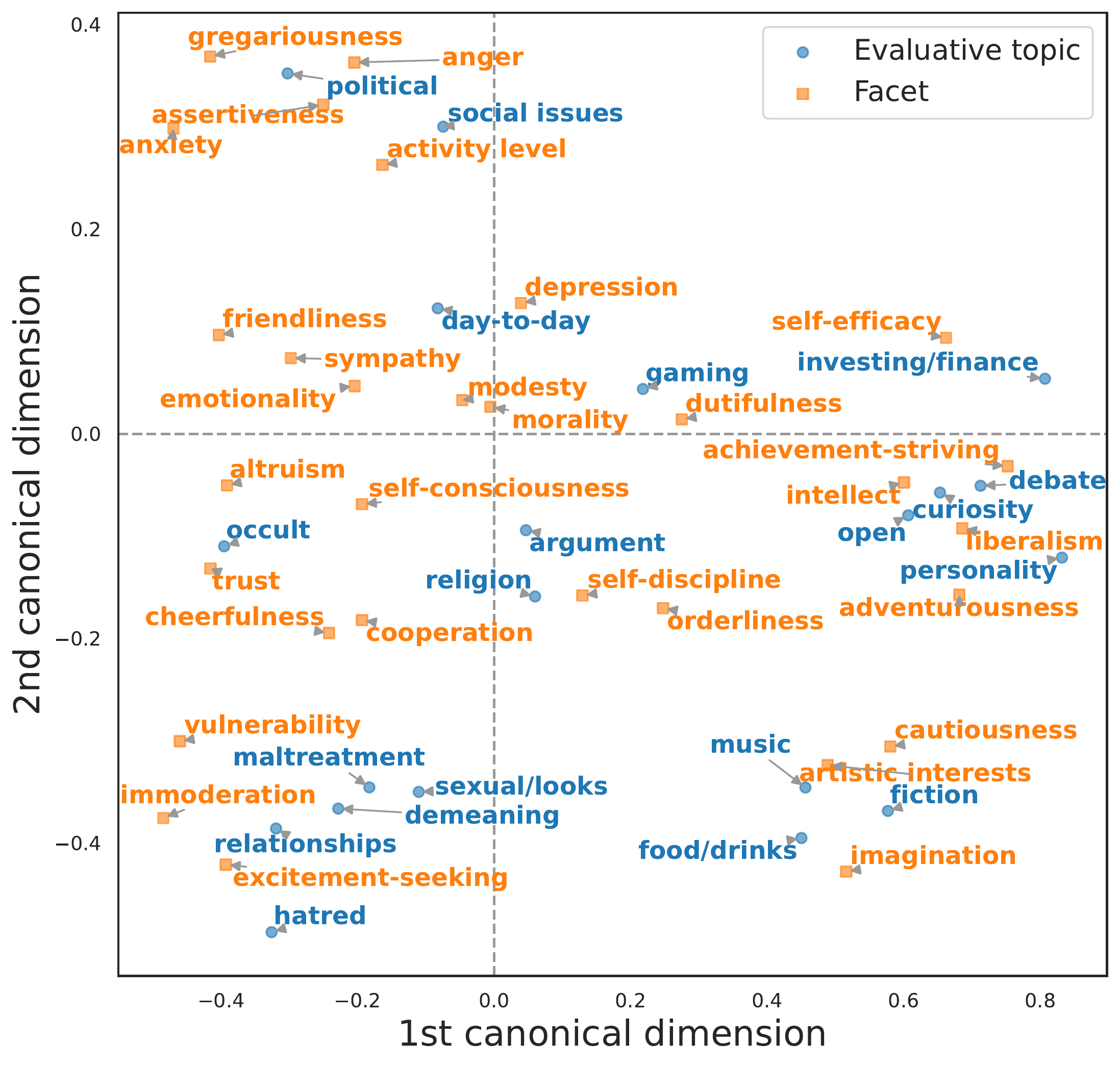}
\caption{Canonical loadings of the first two dimensions for facets and evaluative topics. The proximity of individual points indicates the strength of association between two data points. Specifically, when two points, regardless of their type (topic or facet), have similar angles with respect to the origin and similar magnitudes, this hints that the two points contribute similarly to the canonical variate and have stronger associations.}
\label{fig:ccaproj}
\end{figure}

To further investigate the relationship between facets and evaluative topics as well as their individual importance, we computed canonical loadings, i.e., canonical structure correlations, by projecting both sets of original variables onto the first and second canonical dimensions (Figure \ref{fig:ccaproj}). Canonical loadings reflect the shared variance of the observed variable and the canonical variate. We followed the common practice in psychometrics \citep{braak-1990-interpreting} and computed the intra-set variable-to-variate correlation for the instrument variable (evaluative topic), and the inter-set correlation for the goal variable (facet). We identified three distinct clusters of facets and topics. The first one is at the bottom right corner of \Cref{fig:ccaproj} (\textit{music}, \textit{fiction}; \textit{artistic-interests}, \textit{imagination}) and it indicates the openness aspect of openness to experiences. The bottom-left cluster (\textit{maltreatment}, \textit{demeaning}; \textit{vulnerability}, \textit{immoderation}, and \textit{excitement-seeking}) roughly corresponds to unpleasantly emotionally charged cluster. Finally, the top left cluster  (\textit{political}, \textit{social issues}; \textit{activity level}, \textit{assertiveness}) can be interpreted as social engagement.
We further validate the CCA results by showing the dispersion of facets in the first and second canonical dimensions (\Cref{fig:facetvar1,fig:facetvar2} in the Appendix). We find that facets from the same domain are grouped but that there are also associations between facets from different domains, which is expected based on prior research \cite{schwaba2020facet}.

\section{Conclusion}
\label{sec:conclusion}

The relationship between evaluative language and personality has been understudied in NLP research. We aim to fill this gap by proposing evaluative author profiling -- linking text of authors to topics obtained from text filtered for evaluative language. We applied evaluative profiling on a dataset of Reddit comments with self-reported Big 5 personality facet scores. Using canonical correlation analysis, we showed that the facets within the same trait have stronger associations in the canonical space. We found that evaluative topics have moderate correlations with Big 5 facets. Moreover, we corroborate the hypothesis that evaluative expressions hold greater informational value for personality analysis than unfiltered texts. Additionally, we showed that non-evaluative text has a much weaker association with Big 5 facets compared to evaluative text. Finally, we observed correlations consistent with previous research in personality psychology. We believe that our study can contribute to a better understanding of evaluative language on social media and how it relates to personality traits.

\section*{Ethical Considerations}

Our research has been approved by an academic IRB.

\paragraph{Potential harm.}
Our study is an exploratory study, hence one cannot generalize based on the correlations we obtained. Otherwise, that could lead to unsupported generalization, which may discriminate against certain groups of people.

\paragraph{Collecting data from users.}
According to the American Psychological Association's Ethical Principles,\footnote{https://www.apa.org/ethics/code} researchers may waive informed consent when analyzing archival data (i.e., data collected before the study began) if disclosure of responses would not expose participants to risk of criminal or civil liability or harm.
In our case, we use Reddit data, where Reddit users agree via Reddit User Agreement that they will not disclose any sensitive information to others, and they are informed that their comments are publicly available. As users may opt-out and delete their data, we removed deleted user accounts and comments. We also present our results at the group level rather than the individual level to further protect users' identities.

\paragraph{Misuse potential.}
In principle, evaluative profiling could be used for making decisions at the individual level based on social media textual data, e.g., in micro-targeting. We strongly advocate against the use of our methods for these or other ethically questionable applications.

\paragraph{Biases.}
We note that we used self-reported data from Reddit. As such, this data may not be perfectly accurate and may include various biases, notably the acquisition bias. The dataset we use may not be representative of the Reddit population, and it is certainly not representative of any wider population of users.
\section*{Limitations}

Although our evaluative filtering technique resulted with evaluative topics that have stronger correlations with Big 5 facets than non-evaluative text, filtering cannot be validated in isolation without additional annotation. Additionally, incorporating additional information such as sentiment polarity can hurt interpretability of the results. A possible solution to circumvent this problem is to use structural topic models  \cite{roberts-etal-2014-structural}, which can model additional information such as user demographics as covariates. Since we use topic modeling and try to optimize topic diversity and coherence, which has been critized as of late \cite{holye-etal-2021-automated}, it can be hard to choose appropriate representants for the topic and there are no guarantees that a certain topic will be coherent enough to translate to a meaningful concept. From the psychology point of view, we conduct only an exploratory study where we observe correlation, so we cannot make any confirmation of the results from personality psychology, but only support it with associations. Moreover, since we conduct a facet-level study and take only the scores from the same questionnaire to mitigate the noise from different tests, the number of reports of facet scores is relatively small ($n = 127$) compared to the total number of users with reported traits scores ($n = 1,608$).

Our study was limited to English texts. A potential transfer to other languages would require additional resources, namely adequate lexicons and sentiment classifier to enable evalautive filtering, making the transfer non-trivial. Finally, we would need to retrain the topic models on texts in the target languages.

\section*{Acknowledgements}

We thank the reviewers for their comments. We also thank Irina Masnikosa for her invaluable linguistic advice. This work has been fully supported by the Croatian Science Foundation under the project IP-2020-02-8671 PSYTXT (``Computational Models for Text-Based Personality Prediction and Analysis'').

\bibliography{references/anthology, references/custom}

\begin{thebibliography}{60}
\expandafter\ifx\csname natexlab\endcsname\relax\def\natexlab#1{#1}\fi

\bibitem[{Ackerman and Heggestad(1997)}]{ackerman1997intelligence}
Phillip~L Ackerman and Eric~D Heggestad. 1997.
\newblock Intelligence, personality, and interests: evidence for overlapping
  traits.
\newblock \emph{Psychological bulletin}, 121(2):219.

\bibitem[{Allan(2007)}]{allan2007western}
Keith Allan. 2007.
\newblock \emph{The western classical tradition in linguistics}.
\newblock Equinox London.

\bibitem[{Benamara et~al.(2017)Benamara, Taboada, and
  Mathieu}]{benamara-etal-2017-evaluative}
Farah Benamara, Maite Taboada, and Yannick Mathieu. 2017.
\newblock \href {https://doi.org/10.1162/COLI_a_00278} {Evaluative language
  beyond bags of words: Linguistic insights and computational applications}.
\newblock \emph{Computational Linguistics}, 43(1):201--264.

\bibitem[{Bianchi et~al.(2021)Bianchi, Terragni, and
  Hovy}]{bianchi-etal-2021-pre}
Federico Bianchi, Silvia Terragni, and Dirk Hovy. 2021.
\newblock \href {https://doi.org/10.18653/v1/2021.acl-short.96} {Pre-training
  is a hot topic: Contextualized document embeddings improve topic coherence}.
\newblock In \emph{Proceedings of the 59th Annual Meeting of the Association
  for Computational Linguistics and the 11th International Joint Conference on
  Natural Language Processing (Volume 2: Short Papers)}, pages 759--766,
  Online. Association for Computational Linguistics.

\bibitem[{Biber and Finegan(1989)}]{biberfinegan}
Douglas Biber and Edward Finegan. 1989.
\newblock \href {https://doi.org/doi:10.1515/text.1.1989.9.1.93} {Styles of
  stance in english: Lexical and grammatical marking of evidentiality and
  affect}.
\newblock \emph{Text - Interdisciplinary Journal for the Study of Discourse},
  9(1):93--124.

\bibitem[{Blei and Lafferty(2009)}]{blei-etal-2009-topic}
David~M Blei and John~D Lafferty. 2009.
\newblock Topic models.
\newblock In \emph{Text mining}, pages 101--124. Chapman and Hall/CRC.

\bibitem[{Blei et~al.(2003)Blei, Ng, and Jordan}]{blei2003latent}
David~M. Blei, Andrew~Y. Ng, and Michael~I. Jordan. 2003.
\newblock Latent dirichlet allocation.
\newblock \emph{J. Mach. Learn. Res.}, 3:993–1022.

\bibitem[{Bosco et~al.(2015)Bosco, Singh, Field, and
  Pierce}]{bosco-et-al-2015-correlational}
Frank Bosco, Kulraj Singh, James Field, and Charles Pierce. 2015.
\newblock \href {https://doi.org/10.1037/a0038047} {Correlational effect size
  benchmarks}.
\newblock \emph{Journal of Applied Psychology}, 100:431--449.

\bibitem[{Chauvin and Mullet(2021)}]{chauvin2021individual}
Bruno Chauvin and Etienne Mullet. 2021.
\newblock Individual differences in paranormal beliefs: The differential role
  of personality aspects.
\newblock \emph{Current psychology}, 40(3):1218--1227.

\bibitem[{Cheng et~al.(2014)Cheng, Yan, Lan, and Guo}]{cheng2014btm}
Xueqi Cheng, Xiaohui Yan, Yanyan Lan, and Jiafeng Guo. 2014.
\newblock \href {https://doi.org/10.1109/TKDE.2014.2313872} {Btm: Topic
  modeling over short texts}.
\newblock \emph{IEEE Transactions on Knowledge and Data Engineering},
  26(12):2928--2941.

\bibitem[{Church et~al.(2008)Church, Katigbak, Reyes, Salanga, Miramontes, and
  Adams}]{church2008prediction}
A~Timothy Church, Marcia~S Katigbak, Jose Alberto~S Reyes, Maria Guadalupe~C
  Salanga, Lilia~A Miramontes, and Nerissa~B Adams. 2008.
\newblock Prediction and cross-situational consistency of daily behavior across
  cultures: Testing trait and cultural psychology perspectives.
\newblock \emph{Journal of Research in Personality}, 42(5):1199--1215.

\bibitem[{Costa~Jr and McCrae(1995)}]{costa-etal-1995-domains}
Paul~T Costa~Jr and Robert~R McCrae. 1995.
\newblock Domains and facets: Hierarchical personality assessment using the
  revised neo personality inventory.
\newblock \emph{Journal of personality assessment}, 64(1):21--50.

\bibitem[{Ding et~al.(2018)Ding, Nallapati, and
  Xiang}]{ding-etal-2018-coherence}
Ran Ding, Ramesh Nallapati, and Bing Xiang. 2018.
\newblock \href {https://doi.org/10.18653/v1/D18-1096} {Coherence-aware neural
  topic modeling}.
\newblock In \emph{Proceedings of the 2018 Conference on Empirical Methods in
  Natural Language Processing}, pages 830--836, Brussels, Belgium. Association
  for Computational Linguistics.

\bibitem[{Du et~al.(2017)Du, Xu, He, and Gui}]{du2017stance}
Jiachen Du, Ruifeng Xu, Yulan He, and Lin Gui. 2017.
\newblock \href {https://doi.org/10.24963/ijcai.2017/557} {Stance
  classification with target-specific neural attention}.
\newblock In \emph{Proceedings of the Twenty-Sixth International Joint
  Conference on Artificial Intelligence, {IJCAI-17}}, pages 3988--3994.

\bibitem[{Funder(2012)}]{funder2012accurate}
David~C Funder. 2012.
\newblock Accurate personality judgment.
\newblock \emph{Current Directions in Psychological Science}, 21(3):177--182.

\bibitem[{Gignac and Szodorai(2016)}]{gignac2016effect}
Gilles~E Gignac and Eva~T Szodorai. 2016.
\newblock Effect size guidelines for individual differences researchers.
\newblock \emph{Personality and individual differences}, 102:74--78.

\bibitem[{Gjurkovi{\'c} et~al.(2021)Gjurkovi{\'c}, Karan, Vukojevi{\'c},
  Bo{\v{s}}njak, and {\v{S}}najder}]{gjurkovic2021pandora}
Matej Gjurkovi{\'c}, Mladen Karan, Iva Vukojevi{\'c}, Mihaela Bo{\v{s}}njak,
  and Jan {\v{S}}najder. 2021.
\newblock \href {https://doi.org/10.18653/v1/2021.socialnlp-1.12} {{PANDORA}
  talks: Personality and demographics on {R}eddit}.
\newblock In \emph{Proceedings of the Ninth International Workshop on Natural
  Language Processing for Social Media}, pages 138--152, Online. Association
  for Computational Linguistics.

\bibitem[{Goldberg(1981)}]{goldberg-1981-language}
Lewis~R Goldberg. 1981.
\newblock Language and individual differences: The search for universals in
  personality lexicons.
\newblock \emph{Review of personality and social psychology}, 2(1):141--165.

\bibitem[{He et~al.(2017)He, Lee, Ng, and
  Dahlmeier}]{he-etal-2017-unsupervised}
Ruidan He, Wee~Sun Lee, Hwee~Tou Ng, and Daniel Dahlmeier. 2017.
\newblock \href {https://doi.org/10.18653/v1/P17-1036} {An unsupervised neural
  attention model for aspect extraction}.
\newblock In \emph{Proceedings of the 55th Annual Meeting of the Association
  for Computational Linguistics (Volume 1: Long Papers)}, pages 388--397,
  Vancouver, Canada. Association for Computational Linguistics.

\bibitem[{Henze and Zirkler(1990)}]{henze-zirkler}
Norbert Henze and Bernd Zirkler. 1990.
\newblock A class of invariant consistent tests for multivariate normality.
\newblock \emph{Communications in Statistics-theory and Methods},
  19:3595--3617.

\bibitem[{Hoang et~al.(2019)Hoang, Bihorac, and
  Rouces}]{hoang-etal-2019-aspect}
Mickel Hoang, Oskar~Alija Bihorac, and Jacobo Rouces. 2019.
\newblock \href {https://aclanthology.org/W19-6120} {Aspect-based sentiment
  analysis using {BERT}}.
\newblock In \emph{Proceedings of the 22nd Nordic Conference on Computational
  Linguistics}, pages 187--196, Turku, Finland. Link{\"o}ping University
  Electronic Press.

\bibitem[{Hong and Davison(2010)}]{hong2010empirical}
Liangjie Hong and Brian~D Davison. 2010.
\newblock Empirical study of topic modeling in twitter.
\newblock In \emph{Proceedings of the first workshop on social media
  analytics}, pages 80--88.

\bibitem[{Hoyle et~al.(2021)Hoyle, Goel, Hian-Cheong, Peskov, Boyd-Graber, and
  Resnik}]{holye-etal-2021-automated}
Alexander Hoyle, Pranav Goel, Andrew Hian-Cheong, Denis Peskov, Jordan
  Boyd-Graber, and Philip Resnik. 2021.
\newblock \href
  {https://proceedings.neurips.cc/paper/2021/file/0f83556a305d789b1d71815e8ea4f4b0-Paper.pdf}
  {Is automated topic model evaluation broken? the incoherence of coherence}.
\newblock In \emph{Advances in Neural Information Processing Systems},
  volume~34, pages 2018--2033. Curran Associates, Inc.

\bibitem[{Hu and Liu(2004)}]{hu2004mining}
Minqing Hu and Bing Liu. 2004.
\newblock Mining and summarizing customer reviews.
\newblock In \emph{Proceedings of the tenth ACM SIGKDD international conference
  on Knowledge discovery and data mining}, pages 168--177.

\bibitem[{Hunston(2010)}]{hunston2010corpus}
Susan Hunston. 2010.
\newblock \emph{Corpus approaches to evaluation: Phraseology and evaluative
  language}, volume~13.
\newblock Routledge.

\bibitem[{Hunston and Thompson(2000)}]{hunston-thompson-2000-evaluation}
Susan Hunston and Geoffrey Thompson. 2000.
\newblock \emph{Evaluation in text: Authorial stance and the construction of
  discourse: Authorial stance and the construction of discourse}.
\newblock Oxford University Press, UK.

\bibitem[{Hutto and Gilbert(2014)}]{hutto-gilbert-2014-vader}
Clayton Hutto and Eric Gilbert. 2014.
\newblock Vader: A parsimonious rule-based model for sentiment analysis of
  social media text.
\newblock In \emph{Proceedings of the International AAAI Conference on Web and
  Social Media}, volume~8.

\bibitem[{Hyman(1957)}]{hyman-1957-exploration}
Herbert~H Hyman. 1957.
\newblock An exploration into opinions and personality.
\newblock \emph{World Politics}, 10(1):144--153.

\bibitem[{Intiful et~al.(2019)Intiful, Oddam, Kretchy, and
  Quampah}]{intiful2019exploring}
Freda~Dzifa Intiful, Emefa~Gifty Oddam, Irene Kretchy, and Joana Quampah. 2019.
\newblock Exploring the relationship between the big five personality
  characteristics and dietary habits among students in a ghanaian university.
\newblock \emph{BMC psychology}, 7(1):1--7.

\bibitem[{Kiesling et~al.(2018)Kiesling, Pavalanathan, Fitzpatrick, Han, and
  Eisenstein}]{kiesling-etal-2018-interactional}
Scott~F. Kiesling, Umashanthi Pavalanathan, Jim Fitzpatrick, Xiaochuang Han,
  and Jacob Eisenstein. 2018.
\newblock \href {https://doi.org/10.1162/coli_a_00334} {Interactional
  stancetaking in online forums}.
\newblock \emph{Computational Linguistics}, 44(4):683--718.

\bibitem[{Kulkarni et~al.(2018)Kulkarni, Kern, Stillwell, Kosinski, Matz,
  Ungar, Skiena, and Schwartz}]{kulkarni2018latent}
Vivek Kulkarni, Margaret~L. Kern, David Stillwell, Michal Kosinski, Sandra
  Matz, Lyle Ungar, Steven Skiena, and H.~Andrew Schwartz. 2018.
\newblock \href {https://doi.org/10.1371/journal.pone.0201703} {Latent human
  traits in the language of social media: An open-vocabulary approach}.
\newblock \emph{PLOS ONE}, 13(11):1--18.

\bibitem[{Larson et~al.(2002)Larson, Rottinghaus, and
  Borgen}]{larson2002interests}
Lisa~M. Larson, Patrick~J. Rottinghaus, and Fred~H. Borgen. 2002.
\newblock \href {https://doi.org/https://doi.org/10.1006/jvbe.2001.1854}
  {Meta-analyses of big six interests and big five personality factors}.
\newblock \emph{Journal of Vocational Behavior}, 61(2):217--239.

\bibitem[{Lee and Ma(2012)}]{lee2012news}
Chei~Sian Lee and Long Ma. 2012.
\newblock News sharing in social media: The effect of gratifications and prior
  experience.
\newblock \emph{Computers in human behavior}, 28(2):331--339.

\bibitem[{Luo et~al.(2019)Luo, Ao, Song, Li, Yang, He, and Yu}]{ijcai2019-712}
Ling Luo, Xiang Ao, Yan Song, Jinyao Li, Xiaopeng Yang, Qing He, and Dong Yu.
  2019.
\newblock \href {https://doi.org/10.24963/ijcai.2019/712} {Unsupervised neural
  aspect extraction with sememes}.
\newblock In \emph{Proceedings of the Twenty-Eighth International Joint
  Conference on Artificial Intelligence, {IJCAI-19}}, pages 5123--5129.
  International Joint Conferences on Artificial Intelligence Organization.

\bibitem[{Malrieu(1999)}]{malrieu1999evaluative}
Jean~Pierre Malrieu. 1999.
\newblock \emph{Evaluative semantics: Language, cognition, and ideology}.
\newblock Psychology Press.

\bibitem[{Marshall et~al.(2015)Marshall, Lefringhausen, and
  Ferenczi}]{marshall2015big}
Tara~C Marshall, Katharina Lefringhausen, and Nelli Ferenczi. 2015.
\newblock The big five, self-esteem, and narcissism as predictors of the topics
  people write about in facebook status updates.
\newblock \emph{Personality and Individual Differences}, 85:35--40.

\bibitem[{Mikolov et~al.(2013)Mikolov, Chen, Corrado, and
  Dean}]{mikolov2013efficient}
Tomas Mikolov, Kai Chen, Greg Corrado, and Jeffrey Dean. 2013.
\newblock Efficient estimation of word representations in vector space.
\newblock \emph{arXiv preprint arXiv:1301.3781}.

\bibitem[{Ozer and Benet-Martinez(2006)}]{ozer2006personality}
Daniel~J Ozer and Veronica Benet-Martinez. 2006.
\newblock Personality and the prediction of consequential outcomes.
\newblock \emph{Annual review of psychology}, 57:401.

\bibitem[{Pang and Lee(2008)}]{bo-lee-2008-opinion}
Bo~Pang and Lillian Lee. 2008.
\newblock \href {https://doi.org/10.1561/1500000011} {Opinion mining and
  sentiment analysis}.
\newblock \emph{Found. Trends Inf. Retr.}, 2(1–2):1–135.

\bibitem[{Park et~al.(2014)Park, Schwartz, Eichstaedt, Kern, Kosinski,
  Stillwell, Ungar, and Seligman}]{park-etal-2014-automatic}
Gregory Park, H.~Schwartz, Johannes Eichstaedt, Margaret Kern, Michal Kosinski,
  David Stillwell, Lyle Ungar, and Martin Seligman. 2014.
\newblock \href {https://doi.org/10.1037/pspp0000020} {Automatic personality
  assessment through social media language}.
\newblock \emph{Journal of personality and social psychology}, 108.

\bibitem[{Pavalanathan et~al.(2017)Pavalanathan, Fitzpatrick, Kiesling, and
  Eisenstein}]{pavalanathan-etal-2017-multidimensional}
Umashanthi Pavalanathan, Jim Fitzpatrick, Scott Kiesling, and Jacob Eisenstein.
  2017.
\newblock \href {https://doi.org/10.18653/v1/P17-1082} {A multidimensional
  lexicon for interpersonal stancetaking}.
\newblock In \emph{Proceedings of the 55th Annual Meeting of the Association
  for Computational Linguistics (Volume 1: Long Papers)}, pages 884--895,
  Vancouver, Canada. Association for Computational Linguistics.

\bibitem[{Pennebaker and King(2000)}]{pennebaker-king-2000-linguistic}
James Pennebaker and Laura King. 2000.
\newblock \href {https://doi.org/10.1037//0022-3514.77.6.1296} {Linguistic
  styles: Language use as an individual difference}.
\newblock \emph{Journal of personality and social psychology}, 77:1296--312.

\bibitem[{Pontiki et~al.(2014)Pontiki, Galanis, Pavlopoulos, Papageorgiou,
  Androutsopoulos, and Manandhar}]{pontiki-etal-2014-semeval}
Maria Pontiki, Dimitris Galanis, John Pavlopoulos, Harris Papageorgiou, Ion
  Androutsopoulos, and Suresh Manandhar. 2014.
\newblock \href {https://doi.org/10.3115/v1/S14-2004} {{S}em{E}val-2014 task 4:
  Aspect based sentiment analysis}.
\newblock In \emph{Proceedings of the 8th International Workshop on Semantic
  Evaluation ({S}em{E}val 2014)}, pages 27--35, Dublin, Ireland. Association
  for Computational Linguistics.

\bibitem[{Reimers and Gurevych(2019)}]{reimers-gurevych-2019-sentence}
Nils Reimers and Iryna Gurevych. 2019.
\newblock \href {https://doi.org/10.18653/v1/D19-1410} {Sentence-{BERT}:
  Sentence embeddings using {S}iamese {BERT}-networks}.
\newblock In \emph{Proceedings of the 2019 Conference on Empirical Methods in
  Natural Language Processing and the 9th International Joint Conference on
  Natural Language Processing (EMNLP-IJCNLP)}, pages 3982--3992, Hong Kong,
  China. Association for Computational Linguistics.

\bibitem[{Reimers et~al.(2019)Reimers, Schiller, Beck, Daxenberger, Stab, and
  Gurevych}]{reimers-etal-2019-classification}
Nils Reimers, Benjamin Schiller, Tilman Beck, Johannes Daxenberger, Christian
  Stab, and Iryna Gurevych. 2019.
\newblock \href {https://doi.org/10.18653/v1/P19-1054} {Classification and
  clustering of arguments with contextualized word embeddings}.
\newblock In \emph{Proceedings of the 57th Annual Meeting of the Association
  for Computational Linguistics}, pages 567--578, Florence, Italy. Association
  for Computational Linguistics.

\bibitem[{Roberts et~al.(2014)Roberts, Stewart, Airoldi, Benoit, Blei, Brandt,
  and Spirling}]{roberts-etal-2014-structural}
Margaret~E Roberts, Brandon~M Stewart, Edoardo~M Airoldi, K~Benoit, D~Blei,
  P~Brandt, and A~Spirling. 2014.
\newblock Structural topic models.
\newblock \emph{Retrieved May}, 30:2014.

\bibitem[{Russo et~al.(2015)Russo, Caselli, and
  Strapparava}]{russo-etal-2015-semeval}
Irene Russo, Tommaso Caselli, and Carlo Strapparava. 2015.
\newblock \href {https://doi.org/10.18653/v1/S15-2077} {{S}em{E}val-2015 task
  9: {CLIPE}val implicit polarity of events}.
\newblock In \emph{Proceedings of the 9th International Workshop on Semantic
  Evaluation ({S}em{E}val 2015)}, pages 443--450, Denver, Colorado. Association
  for Computational Linguistics.

\bibitem[{Schwaba et~al.(2020)Schwaba, Rhemtulla, Hopwood, and
  Bleidorn}]{schwaba2020facet}
Ted Schwaba, Mijke Rhemtulla, Christopher~J Hopwood, and Wiebke Bleidorn. 2020.
\newblock A facet atlas: Visualizing networks that describe the blends, cores,
  and peripheries of personality structure.
\newblock \emph{PloS one}, 15(7):e0236893.

\bibitem[{Schwartz et~al.(2013)Schwartz, Eichstaedt, Kern, Dziurzynski,
  Ramones, Agrawal, Shah, Kosinski, Stillwell, Seligman, and
  Ungar}]{schwartz-etal-2013-personality}
H.~Andrew Schwartz, Johannes~C. Eichstaedt, Margaret~L. Kern, Lukasz
  Dziurzynski, Stephanie~M. Ramones, Megha Agrawal, Achal Shah, Michal
  Kosinski, David Stillwell, Martin E.~P. Seligman, and Lyle~H. Ungar. 2013.
\newblock \href {https://doi.org/10.1371/journal.pone.0073791} {Personality,
  gender, and age in the language of social media: The open-vocabulary
  approach}.
\newblock \emph{PLOS ONE}, 8(9):1--16.

\bibitem[{Skimina et~al.(2021)Skimina, Cieciuch, and Strus}]{skimina2021traits}
Ewa Skimina, Jan Cieciuch, and W{\l}odzimierz Strus. 2021.
\newblock Traits and values as predictors of the frequency of everyday
  behavior: Comparison between models and levels.
\newblock \emph{Current Psychology}, 40(1):133--153.

\bibitem[{Soto(2019)}]{soto2019replicable}
Christopher~J Soto. 2019.
\newblock How replicable are links between personality traits and consequential
  life outcomes? the life outcomes of personality replication project.
\newblock \emph{Psychological Science}, 30(5):711--727.

\bibitem[{Stewart et~al.(2022)Stewart, M{\~o}ttus, Seeboth, Soto, and
  Johnson}]{stewart2022finer}
Ross~David Stewart, Ren{\'e} M{\~o}ttus, Anne Seeboth, Christopher~John Soto,
  and Wendy Johnson. 2022.
\newblock The finer details? the predictability of life outcomes from big five
  domains, facets, and nuances.
\newblock \emph{Journal of personality}, 90(2):167--182.

\bibitem[{ter Braak(1990)}]{braak-1990-interpreting}
Cajo ter Braak. 1990.
\newblock \href {https://doi.org/10.1007/BF02294765} {Interpreting canonical
  correlation analysis through biplots of stucture correlations and weights}.
\newblock \emph{Psychometrika}, 55:519--531.

\bibitem[{Terragni et~al.(2021)Terragni, Fersini, and
  Messina}]{terragni-etal-2021-natural}
Silvia Terragni, Elisabetta Fersini, and Enza Messina. 2021.
\newblock \href {https://doi.org/10.1007/978-3-030-80599-9_4} {Word
  embedding-based topic similarity measures}.
\newblock In \emph{Natural Language Processing and Information Systems: 26th
  International Conference on Applications of Natural Language to Information
  Systems, NLDB 2021, Saarbr\"{u}cken, Germany, June 23–25, 2021,
  Proceedings}, page 33–45, Berlin, Heidelberg. Springer-Verlag.

\bibitem[{Thompson and Hunston(2000)}]{hunston2000evaluation}
Geoffrey Thompson and Susan Hunston. 2000.
\newblock Evaluation: an introduction.
\newblock In \emph{Evaluation in text: Authorial stance and the construction of
  discourse}. Oxford University Press.

\bibitem[{Toh and Wang(2014)}]{toh-wang-2014-dlirec}
Zhiqiang Toh and Wenting Wang. 2014.
\newblock \href {https://doi.org/10.3115/v1/S14-2038} {{DLIREC}: Aspect term
  extraction and term polarity classification system}.
\newblock In \emph{Proceedings of the 8th International Workshop on Semantic
  Evaluation ({S}em{E}val 2014)}, pages 235--240, Dublin, Ireland. Association
  for Computational Linguistics.

\bibitem[{Tulkens and van
  Cranenburgh(2020)}]{tulkens-van-cranenburgh-2020-embarrassingly}
St{\'e}phan Tulkens and Andreas van Cranenburgh. 2020.
\newblock \href {https://doi.org/10.18653/v1/2020.acl-main.290} {Embarrassingly
  simple unsupervised aspect extraction}.
\newblock In \emph{Proceedings of the 58th Annual Meeting of the Association
  for Computational Linguistics}, pages 3182--3187, Online. Association for
  Computational Linguistics.

\bibitem[{Wiebe et~al.(2004)Wiebe, Wilson, Bruce, Bell, and
  Martin}]{wiebe-etal-2004-learning}
Janyce Wiebe, Theresa Wilson, Rebecca Bruce, Matthew Bell, and Melanie Martin.
  2004.
\newblock \href {https://doi.org/10.1162/0891201041850885} {Learning subjective
  language}.
\newblock \emph{Computational Linguistics}, 30(3):277--308.

\bibitem[{Wittmann(2012)}]{wittmann2012principles}
Werner~W Wittmann. 2012.
\newblock Principles of symmetry in evaluation research with implications for
  offender treatment.
\newblock \emph{Antisocial behavior and crime. Contributions of developmental
  and evaluation research to prevention and intervention}, (2011):357--368.

\bibitem[{Zuo et~al.(2016)Zuo, Wu, Zhang, Lin, Wang, Xu, and
  Xiong}]{zuo-etal-2016-topic}
Yuan Zuo, Junjie Wu, Hui Zhang, Hao Lin, Fei Wang, Ke~Xu, and Hui Xiong. 2016.
\newblock \href {https://doi.org/10.1145/2939672.2939880} {Topic modeling of
  short texts: A pseudo-document view}.
\newblock In \emph{Proceedings of the 22nd ACM SIGKDD International Conference
  on Knowledge Discovery and Data Mining}, KDD '16, page 2105–2114, New York,
  NY, USA. Association for Computing Machinery.

\end{thebibliography}
\bibliographystyle{references/acl_natbib}

\clearpage
\appendix

\section{Technical details}
\label{sec:A}

\subsection{Dataset}
\label{subsec:A_dataset}
We selected the comments from users with Big Five self-reports on \textsc{Pandora}. We kept only the comments in English and removed unlikely candidates for evaluative expressions, including comments that were too short -- those with fewer than five words -- and noisy comments -- those consisting of 50\% or more non-alphanumeric characters.
We segmented the comments into sentences using spaCy's English pipeline \textit{en\_core\_web\_lg}\footnote{\href{https://spacy.io/models/en\#en\_core\_web\_lg}{https://spacy.io/models/en\#en\_core\_web\_lg}} to proceed at a finer level. Once again, we discarded texts that were too short, but this time the sentences with fewer than three words.

\subsection{Evaluative filtering}
\label{subsec:A_filtering}

When we employed QSB to mine paraphrases of the initial set of evaluative expressions, we aimed for a tenfold increase in size of the seed set (29k sentences), so we tuned the values of $t_{sim}$ and $\gamma$. We used grid search with the parameter range $t_{sim} \in [0.50, 0.90]$ with step $0.05$ and $\gamma \in [0.90, 1.10]$ with a step of $0.05$ in both cases, which led us to $t_{sim} = 0.7$ and $\gamma = 1.05$. QSB yielded 310k sentences with evaluative markers.

\subsection{Topic modeling}
\label{subsec:A_topic}
We used \textit{scikit-learn} implementation of LDA.\footnote{\href{https://scikit-learn.org/stable/modules/generated/sklearn.decomposition.LatentDirichletAllocation.html}{https://scikit-learn.org/stable/modules/generated/sklearn.decomposition.LatentDirichletAllocation.html}} We adapt the code from the original papers for BTM \cite{cheng2014btm}, ABAE \cite{he-etal-2017-unsupervised}, and CTM \cite{bianchi-etal-2021-pre}. The neural-based architectures in ABAE and CTM have $100,020$ and $19,794,280$, respectively. We trained ABAE for $5$ epochs and CTM for $20$ epochs, with average running times of $71.4$ and $590.8$ minutes over $10$ different runs.  

We used the same preprocessing for all topic models that we experimented with. Specifically, we removed punctuation and lowercased the text. We also eliminated stop words, URLs, emails, digits, and currency symbols, after which we lemmatized the tokens. We repeated each experiment $10$ times with different seeds. We conducted a grid search with the number of topics as a hyperparameter in the $[5, 100]$ range with step $5$. For the rest of the hyperparameters, we used default values.

\subsection{Computing infrastructure}
We conducted our experiments on $2 \times$ \textit{AMD Ryzen Threadripper 3970X 32-Core Processors} and $2 \times$ \textit{NVIDIA GeForce RTX 3090} GPU's with $24$GB of RAM. We used \textit{PyTorch} version $1.9.0$ and CUDA $11.4$.

\section{Additional analysis}
\label{sec:B}
In \Cref{fig:corr}, there are cases where certain evaluative topics show significant correlation with several facets from the same trait. For instance, the topic \textit{social issues} correlates positively and significantly with \textit{anger}, \textit{immoderation}, and \textit{anxiety} facets from the \textit{neuroticism} trait. On the other hand, there are a number of cases where the correlation between facets from the same trait and a topic is of a different sign. This corroborates our hypothesis that topics align better with facets than with traits.

\begin{figure}[p]
\centering
\includegraphics[width=\linewidth]{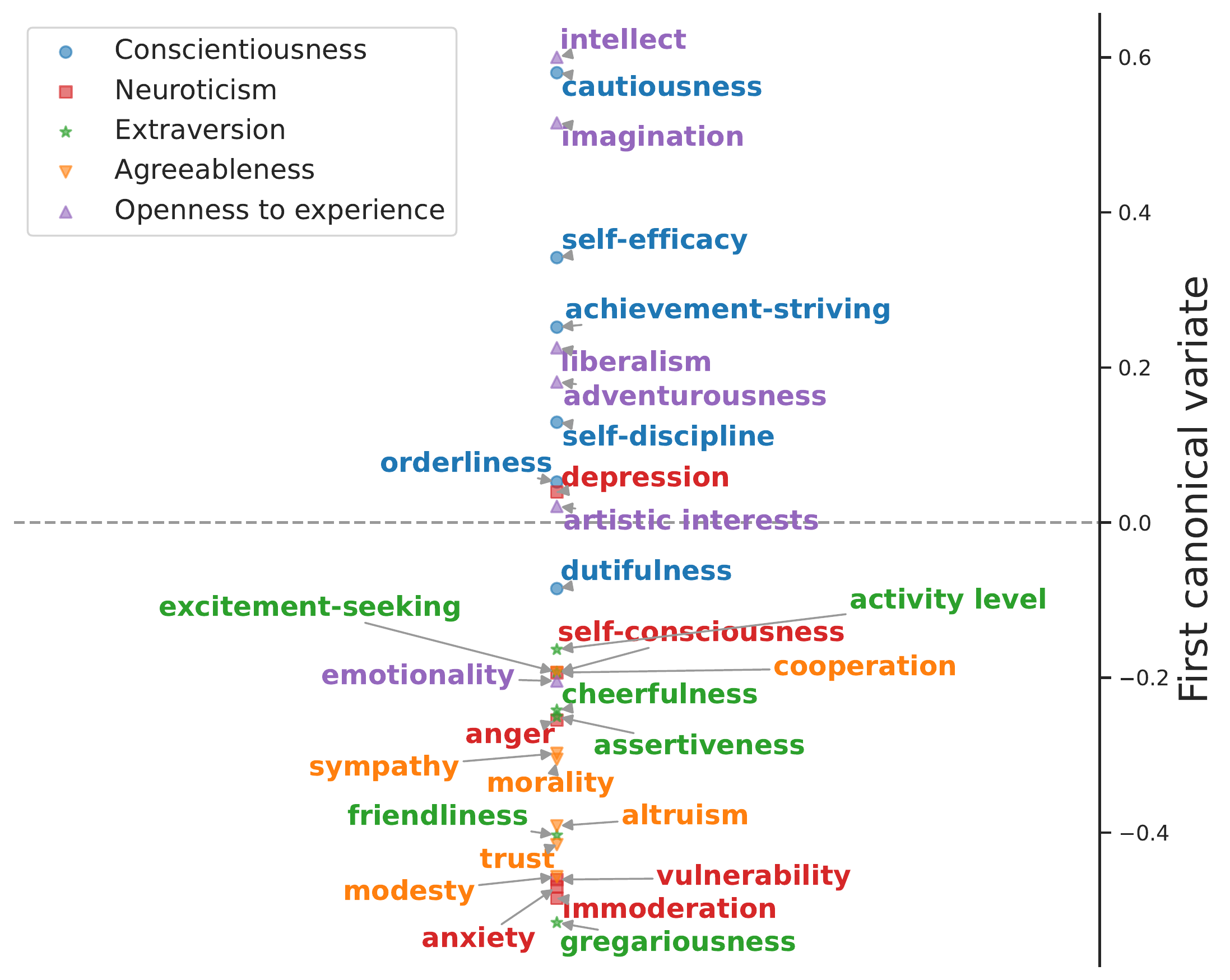}
\caption{Facet projection onto the first canonical variate. The values on the y-axis represent the canonical loadings of facets in the first canonical dimension.}
\label{fig:facetvar1}
\end{figure}

\begin{figure}[p]
\centering
\includegraphics[width=\linewidth]{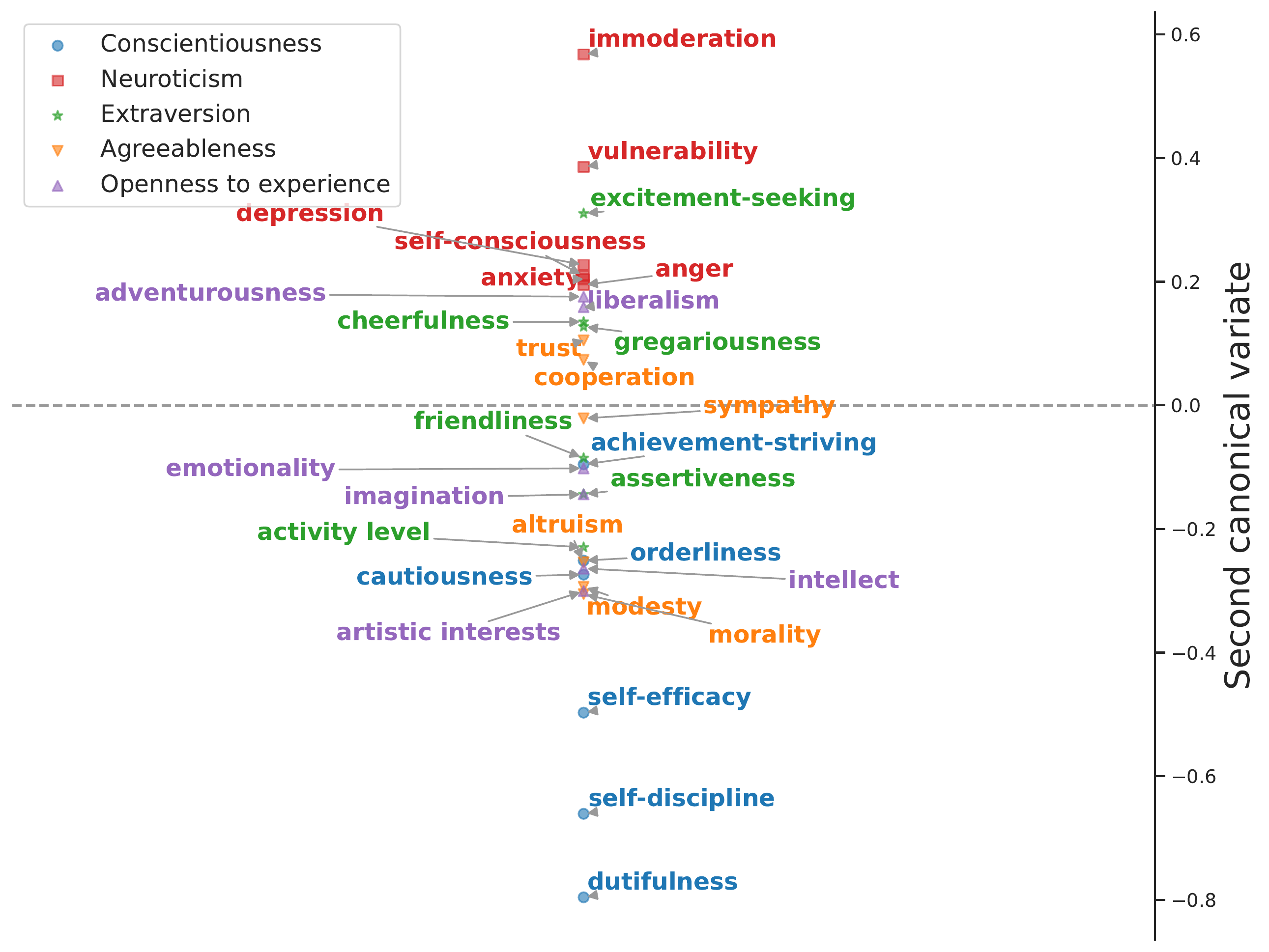}
\caption{Facet projection onto the second canonical variate. The second dimension provides a perfect separation of conscientiousness and neuroticism, with a mixture of facets from the remaining traits.}
\label{fig:facetvar2}
\end{figure}

\end{document}